\documentclass[]{ceurart}

\usepackage{listings}
\usepackage{placeins}
\usepackage{subcaption}
\begin{document}

\copyrightyear{2026}
\copyrightclause{Copyright for this paper by its authors.
  Use permitted under Creative Commons License Attribution 4.0
  International (CC BY 4.0).}

\conference{SLM4ED'26: The 1st Workshop of Small Language Models for Education (SLM4ED),
  June 28, 2026, Seoul, Republic of Korea}

\title{Bounding Boxes to Improve Small Language Model Performance on Vision-Based Grading Tasks}

\author[1]{Lachlan McGinness}[%
email=lachlan.mcginness@anu.edu.au,
]
\address[1]{Australian National University, Acton, ACT, Australia}

\begin{abstract}
  The deployment of Small Language Models (SLMs) in educational settings offers significant advantages in terms of privacy, cost, and scalability.
  However, SLMs often struggle with complex vision-based tasks, such as grading handwritten student exams, due to the high computational cost of processing large images and the visual distractions present on a full page. 
  In this paper, we investigate whether cropping student responses using bounding boxes can improve the accuracy and computational efficiency of SLMs on a short-answer grading task. 
  Using a dataset of scanned handwritten responses from the 2025 Australian Physics Olympiad, we evaluate the performance of several models ranging from 4B to 72B parameters under varying conditions of Chain of Thought (CoT) prompting and image cropping. 
  Our results demonstrate that using bounding boxes significantly improves grading accuracy and reduces computational cost (FLOPs) across models. 
  We conclude that bounding boxes are a crucial pre-processing step for deploying SLMs in large-scale, vision-based educational assessments.
\end{abstract}


\maketitle

\section{Introduction}

The integration of Artificial Intelligence in Education (AIED) has rapidly expanded the use of Large Language Models (LLMs) including Vision Language Models (VLMs) for tasks such as automated short-answer grading \cite{Henkel2025Can,Kortemeyer2023Toward,Kortemeyer2025Assessing,Kortemeyer2024Automated,Liu2024AIassisted,Mok2024Using,Chen2024Achieving,Chen2025Grading,McGinness2025Overview}. 
These studies use large, cloud-hosted, proprietary models.
In parallel, a growing body of work has begun to examine Small Language Models (SLMs)\footnote{typically taken to mean open-weights models with up to about $3\times10^{10}$ parameters} as alternatives that can be deployed locally on consumer-grade hardware \cite{Baumgartner2025Combining,Jaldi2026Small,McGinness2025Can,Reza2025Small}. 
These studies indicate that SLMs can achieve competitive performance on a range of educational generation and evaluation tasks while offering important advantages in privacy, cost, and institutional control over data. 
These properties are particularly attractive in schools, where sensitive student data, regulatory constraints, and limited budgets could prevent the use of proprietary cloud-based LLMs.

A particularly demanding application of language models in education is the grading of \emph{handwritten} student work. 
In contrast to textual short-answer grading, where the student's response is already digitised, handwritten exams require the model to locate the answer on the page, interpret crossed-out text, and ignore unrelated content from other questions. 

VLMs tokenise images by partitioning them into patches that are projected into the language model's embedding space, so that the number of \emph{visual} tokens scales (approximately) linearly with the image area. 
Modern dynamic-resolution architectures such as Gemma3 and Qwen2.5 preserve native aspect ratios and produce visual token counts that are roughly proportional to the number of $14 \times 14$ pixel patches in the input \cite{Gemma3,Yang2025Qwen}.
A full A3 double-page exam scan can therefore consume thousands of visual tokens, and recent analyses have shown that vision-encoder cost grows rapidly with input resolution, often dominating the computational cost of a forward pass \cite{vasu2025fastvlm}. 
Under the standard $C \approx 2nN$ approximation for inference compute \cite{Kaplan2020Scaling,Eimler2026Environmental}, where $n$ is the number of tokens processed and $N$ is the number of active parameters, this directly translates into a higher number of FLOPs (computational cost) per graded response.

VLMs can be easily distracted by content-rich pages which contain a large number of visual features that are irrelevant to any one sub-question; for example other questions or scribbles. 
Therefore SLMs may benefit significantly from reducing the visual search problem before the model sees the image. 
Classical document layout analysis has long relied on cropping a region of interest as a preprocessing step prior to OCR and we hypothesise that selectively decoding only regions relevant to a query will improve both accuracy and efficiency.

In this short study, we test this intuition empirically in an educational setting. 
Using 126 scanned handwritten responses to Question~1a of the 2025 Australian Physics Olympiad, we evaluate eight open-source VLMs ranging from 4B to 72B parameters with two experimental variables: (i) using of CoT prompting and (ii) whether the model is given the full A3 double-page spread or only a bounding-box crop of the relevant question/answer region. 
We measure both grading accuracy against human markers and the total inference FLOPs implied by the input/output token counts.

\section{Methodology}
\label{sec:handwritten_methods}

For our experiments we used Question 1a of the 2025 Australian Physics Olympiad exam. 
Students completed this exam on paper supervised by their teachers. 
As students registered online, they were provided with a participant information sheet and able to tick a box to opt-in to consenting for their responses to be used for the research. 
In total, 588 of the 1357 students who sat the exam consented, and their exams were scanned as double-page spreads (landscape A3 pages).
To reduce the computational cost and time to run this preliminary experiment, 126 of the 588 student exam papers were chosen randomly as the dataset for each of our four experimental conditions. 
Question 1a was a short answer question, worded as follows:

\begin{quotation}
	Imagine a human walking on perfectly flat ground. With each step they accelerate and decelerate. 
	Which force is responsible for this acceleration and deceleration?
\end{quotation}

The answer is simply `friction' or `friction force'. The main challenge for grading this question is reading the student's handwritten response, as very little reasoning is required.

\subsection{Experimental Conditions}

In this experiment, we investigated two sets of independent variables: Chain of Thought (CoT) prompting and bounding boxes. 

\paragraph{Chain of Thought Prompting} 
To investigate the influence of chain of thought prompting, each of the eight models were used to grade the 126 responses with two different prompts. The prompt for the CoT condition was \footnote{Note that although Question 1a is about human walking, the rest of Question 1 is about snail locomotion and there is a large diagram of a snail immediately below the question.}:
\begin{quotation}
	``This answer page contains a student response to a question about snails. 
	Please read the student's answer for Question 1a near the top of the page.
	If they wrote `friction force', `friction', or something similar please output 1.
	Otherwise output 0. If the student wrote friction and then crossed it out and then wrote something else write 0.
	Think this through step by step but then ensure `0' or `1' is the last thing that you write.''
\end{quotation}
\noindent The prompt for the no-CoT condition was:
\begin{quotation}
	This answer page contains a student response to a question about snails. 
	Please read the student's answer for Question 1a near the top of the page.
	If they wrote `friction force', `friction', or something similar please output 1.
	Otherwise output 0. If the student wrote friction and then crossed it out and then wrote something else write 0.
	Please don't write anything else except the 1 or 0 based on the student response. 
	If you do write something else, ensure `0' or `1' is the last thing that you write.
\end{quotation}
The distinguishing factor between these prompts is that in CoT, the model is asked to `think this through step by step'. 
CoT was one of the most successful early prompt engineering techniques \cite{wei2022chain}. 
More recently, Language Models have been trained to natively engage in step by step thinking, so CoT prompts have a lesser \cite{McGinness2025Imitate} or negative effect \cite{Liu2025Mind}.

\paragraph{Bounding Boxes} 
By inspection of the scanned exams, the researchers determined that pixels corresponding to Question 1a were always in a very similar location. 
This allowed the researchers to automatically apply bounding boxes to crop and capture the students' answers to the question. 
In the bounding box condition, only these areas were given to the model to determine the grade. 
This means that there should be fewer distractions for the model when grading the question. 
The bounding boxes were quite generous so that if the students wrote off to the side, their response would still successfully be captured, allowing for small variations in paper position during scanning. 

\begin{figure}[htbp]
	\centering
	\begin{subfigure}[b]{0.4\textwidth}
		\centering
		\includegraphics[width=\textwidth]{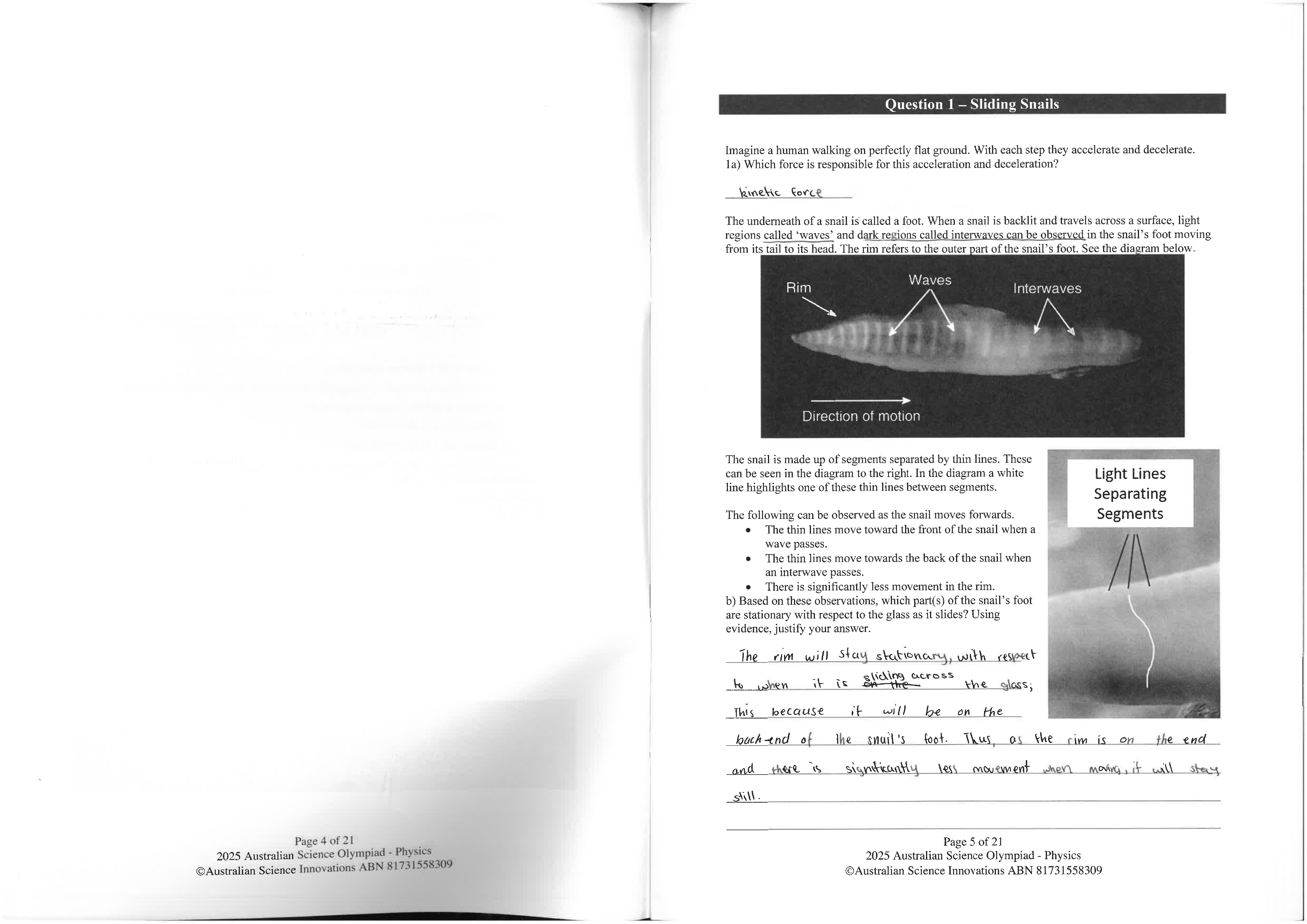}
	\end{subfigure}
	\hfill
	\begin{subfigure}[b]{0.56\textwidth}
		\centering
		\includegraphics[width=\textwidth]{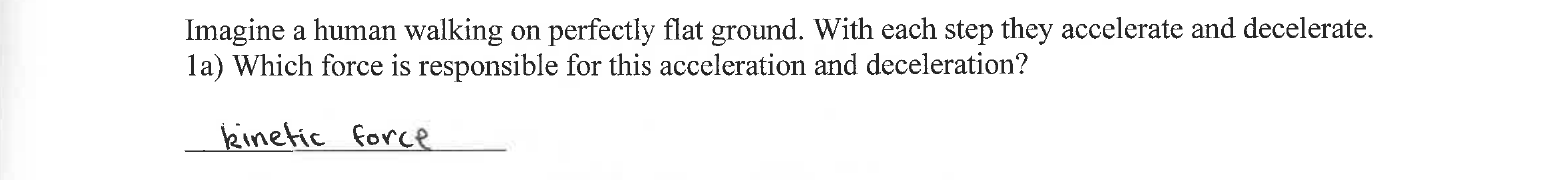}
        \vspace{1.5cm}
	\end{subfigure}
	\caption{Visual comparison of the bounding box and non-bounding box experimental conditions. \textbf{Left:} Original scan of student exam paper. \textbf{Right:} Cropped bounding box of the same student exam paper. The bounding box reduces the number of visual features that could confuse smaller models.}
	\label{fig:boundingbox_example}
\end{figure}

In the non-bounding box condition, the entire double-page spread was provided to the model.
The model was then required to find the student response to question 1a and extract it. 
Figure \ref{fig:boundingbox_example} gives an example of the same student response with and without bounding boxes.

Intuitively, the size of an image is approximately proportional to the number of tokens that it takes to encode it. 
Therefore presenting a model with a smaller picture will require fewer input tokens and reduce the overall computational cost. 

In total, there were four experimental conditions: CoT with bounding boxes, CoT without bounding boxes, no-CoT with bounding boxes, and no-CoT without bounding boxes. 
For each experimental condition, we recorded the raw model response, the number of input tokens, and the number of output tokens. 
The number of tokens was used to calculate the total FLOPs using the $2nN$ approximation. 
Since the model's response is binary ($0$ or $1$), accuracy was determined by calculating the fraction of responses for which the model agrees with the human marking team (which had no discrepancies on this question).

\subsection{Models Evaluated}
The models used in this experiment were open-source and small enough to run on upper-end consumer-grade hardware, making them ideal candidates for local educational deployments. 
The models evaluated are summarised in Table \ref{tab:Bounding_Box_Experiment_Models}.
Note that we included larger models (32B-72B active parameters) as performance ceilings for comparison.

\begin{table}[h]
	\centering
	\caption{Vision-language models evaluated in the bounding box and CoT experiments. For Mixture-of-Experts models, active parameters refer to the number engaged per forward pass.}
	\label{tab:Bounding_Box_Experiment_Models}
	\scalebox{0.8}{
	\begin{tabular}{llccc}
		\toprule
		\textbf{Model} & \textbf{Developer} & \textbf{Active Params} & \textbf{Total Params} & \textbf{Release Date} \\
		\midrule
		Llama 4 Scout        & Meta        & 17B  & 109B & April 2025 \\
		Mistral Small 3.2    & Mistral AI  & 24B  & 24B  & June 2025  \\
		Gemma 3 27B          & Google      & 27B  & 27B  & March 2025 \\
		Gemma 3 12B          & Google      & 12B  & 12B  & March 2025 \\
		Gemma 3 4B           & Google      & 4B   & 4B   & March 2025 \\
		Qwen2.5-VL 72B       & Alibaba     & 72B  & 72B  & January 2025 \\
		Qwen2.5-VL 32B       & Alibaba     & 32B  & 32B  & March 2025 \\
		Qwen2.5-VL 7B        & Alibaba     & 7B   & 7B   & January 2025 \\
		\bottomrule
	\end{tabular}}
\end{table}

\section{Results}

Table \ref{tab:bounding_box_accuracy_results} shows the performance of each of the models in each of the four experimental conditions. Experimental conditions using bounding boxes were the most accurate for all models. 
With the exception of Mistral 3.2, models did not benefit significantly from CoT prompting.
This anomaly may have occurred because Mistral's instruction-following training was significantly different to that of the other models.

\begin{table}[htbp]
	\centering
	\caption{Accuracy for each model across the four experimental conditions. Values reported as mean $\pm$ standard deviation. For most models, the combination of no-CoT prompt with bounding boxes produced the most accurate results.}
	\label{tab:bounding_box_accuracy_results}
	\resizebox{\columnwidth}{!}{%
		\begin{tabular}{lcccccccc}
			\toprule
			\textbf{Condition} & \textbf{Gemma 4B} & \textbf{Gemma 12B} & \textbf{Gemma 27B} & \textbf{Llama4} & \textbf{Mistral3.2} & \textbf{Qwen 7B} & \textbf{Qwen 32B} & \textbf{Qwen 72B} \\
			\midrule
			CoT, No BB & $0.49 \pm 0.04$ & $0.51 \pm 0.04$ & $0.80 \pm 0.04$ & $0.77 \pm 0.04$ & $0.80 \pm 0.04$ & $0.65 \pm 0.04$ & $0.78 \pm 0.04$ & $0.87 \pm 0.03$ \\
			No-CoT, No BB & $0.47 \pm 0.04$ & $0.65 \pm 0.04$ & $0.68 \pm 0.04$ & $0.66 \pm 0.04$ & $0.51 \pm 0.04$ & $0.92 \pm 0.02$ & $0.83 \pm 0.03$ & $\mathbf{0.94 \pm 0.02}$ \\
			CoT + BB & $0.55 \pm 0.04$ & $0.62 \pm 0.04$ & $0.85 \pm 0.03$ & $0.79 \pm 0.04$ & $\mathbf{0.93 \pm 0.02}$ & $0.79 \pm 0.04$ & $0.90 \pm 0.03$ & $\mathbf{0.94 \pm 0.02}$ \\
			No-CoT + BB & $\mathbf{0.56 \pm 0.04}$ & $\mathbf{0.74 \pm 0.04}$ & $\mathbf{0.87 \pm 0.03}$ & $\mathbf{0.90 \pm 0.03}$ & $0.64 \pm 0.04$ & $\mathbf{0.95 \pm 0.02}$ & $\mathbf{0.94 \pm 0.02}$ & $\mathbf{0.94 \pm 0.02}$ \\
			\bottomrule
		\end{tabular}%
	}
\end{table}

\begin{figure}[htbp]
	\centering
	\scalebox{0.8}{
	\includegraphics[width=\textwidth]{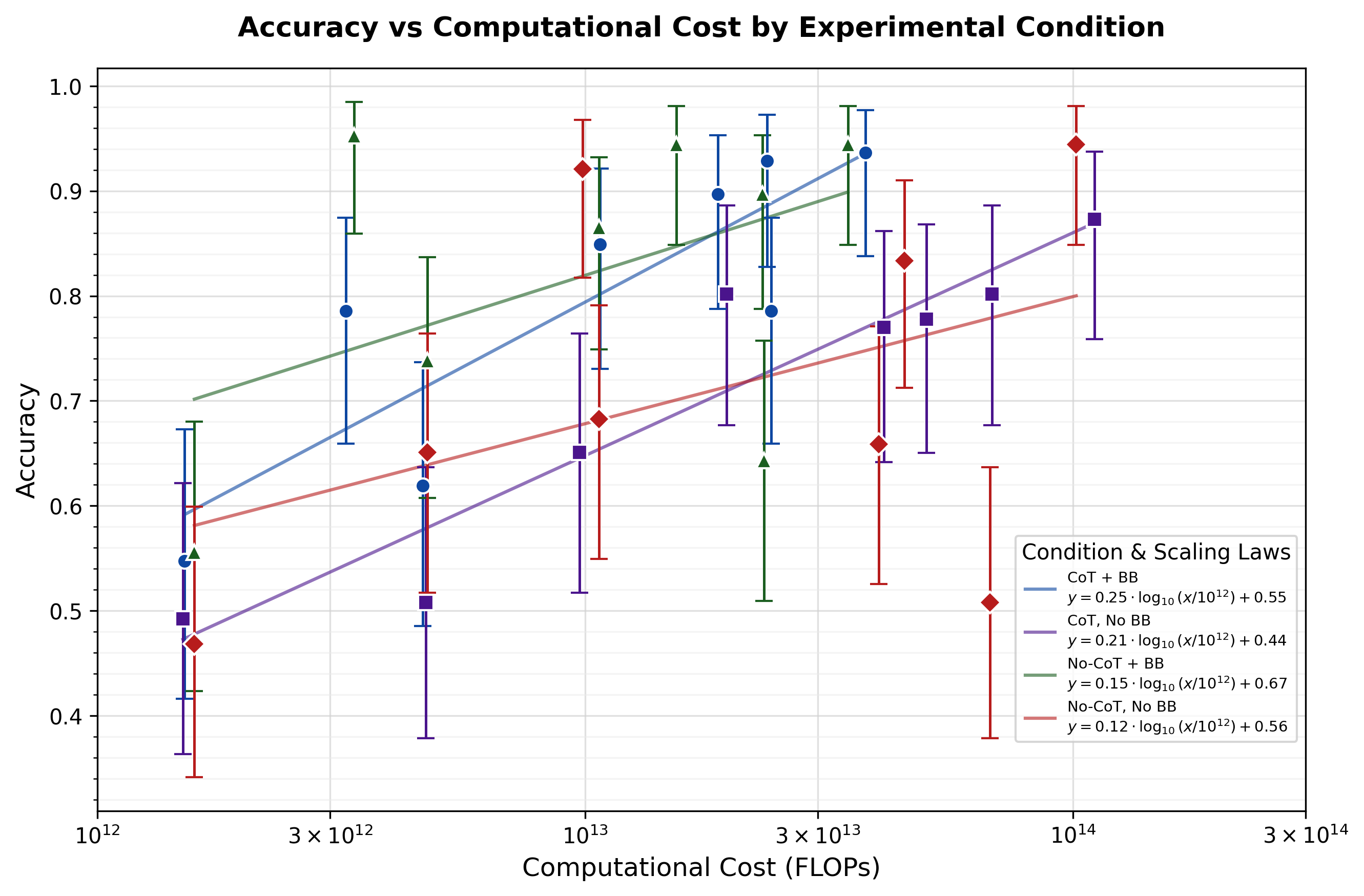}}
	\caption{Plot of accuracy against computational cost for all models and experimental conditions. Error bars indicate the maximum and minimum values of the Wilson Score interval at the Z=3 ($p=$99.7\%) significance level.}
	\label{fig:scatter_plot_bounding_box}
\end{figure}

Figure \ref{fig:scatter_plot_bounding_box} illustrates model performance as a function of computational cost (in FLOPs). 
The data is clearly very noisy as performance varies wildly between models, but there are three trends that emerge.
Firstly, as expected, increased computational expense increases accuracy  across all experimental conditions.
Second, there is little difference between the CoT and no-CoT conditions.
This is displayed in Figure \ref{fig:scatter_plot_bounding_box} as CoT + Bounding Boxes (blue) is similar to No-CoT + Bounding Boxes (green). 
The same pattern applies for the two non-bounding box conditions (purple and red). 
Finally, there is a noticeable improvement of the bounding box conditions (green and blue) over the non-bounding box conditions (red and purple).

The significance of this difference in accuracy across all models can be determined by taking the average and standard error in the mean across all trials as shown in Table \ref{Tab:OverallQ1a}. 

\begin{table}[h]
	\caption{Average performance metrics across all models under different experimental conditions. The uncertainty is the standard error in the mean across all eight models. The reported tokens are the sum of input and output tokens.}
	\label{Tab:OverallQ1a}
	\centering
	\begin{tabular}{lccc}
		\hline
		Experimental Condition & Accuracy & Average Tokens & Average FLOPs (Trillions) \\
		\hline
		CoT, Bounding Box & $0.817 \pm 0.019$ & 694 & 18 \\
		CoT, No Bounding Box & $0.734 \pm 0.019$ & 1549 & 45 \\
		No CoT, Bounding Box & $0.840 \pm 0.020$ & 679 & 17 \\
		No CoT, No Bounding Box & $0.728 \pm 0.024$ & 1463 & 41 \\
		\hline
	\end{tabular}
\end{table}

The accuracy is significantly better for cases where bounding boxes were utilised. The difference between the lowest bounding box accuracy and highest non-bounding box accuracy is $3.08$ times the squared quadrature of the standard errors, corresponding to significance at the 99.8\% confidence level. Furthermore, bounding boxes reduced the number of tokens required by a significant margin compared to parsing the full image, resulting in substantially less computational expense. 

\section{Conclusion}

As educational institutions look to adopt AI, SLMs present a secure, transparent, and cost-effective alternative to large, cloud-based commercial models. 
However, effectively deploying SLMs requires architectural frameworks designed to compensate for their specific limitations. 
Our research demonstrates that when evaluating handwritten exams, providing SLMs with an entire double-page spread degrades their accuracy and unnecessarily increases token processing costs. 

By implementing bounding boxes as a simple preprocessing step, we significantly reduced visual noise and token overhead. 
This intervention improved accuracy and reduced computational cost. 
We recommend that future research agendas focusing on SLMs in education should prioritise removing unnecessary visual clutter using techniques like  bounding boxes, thereby optimising deployment of SLMs in learning analytics and automated grading.

\begin{acknowledgments}
  This research was supported by the authors' respective institutions. We thank the organisers of the Australian Physics Olympiad for providing access to the anonymised student response dataset. 
\end{acknowledgments}

\section*{Declaration on Generative AI}
During the preparation of this work, the author(s) used Large Language Models to assist with literature review, formatting, preparing figures, drafting and proof reading. The author(s) carefully reviewed and edited LLM generated content and take full responsibility for the publication’s content. 

\bibliography{PhDThesisReferences}

\end{document}